\documentclass{article}%
\usepackage{iclr2027_conference,times}

\usepackage{graphicx}
\usepackage{wrapfig}
\usepackage{needspace}
\usepackage{amsmath}
\usepackage{amssymb}
\usepackage{xcolor}
\usepackage{booktabs}
\usepackage{multirow}
\usepackage{tikz}
\usepackage{natbib}
\usepackage{caption}
\usepackage{subcaption}
\usepackage{changepage}
\usepackage{algorithm}
\usepackage{enumitem}
\usepackage{algorithmic}
\usepackage{newfloat}
\usepackage{listings}
\DeclareCaptionStyle{ruled}{labelfont=normalfont,labelsep=colon,strut=off}
\floatstyle{ruled}
\newfloat{listing}{tb}{lst}{}
\floatname{listing}{Listing}

\usepackage{hyperref}
\usepackage{cleveref}
\usepackage{url}
\hypersetup{
  hidelinks,
  pdfauthor={Ruixuan Huang, Yipei Wang, Wenyi Fang, Hantao Huang, Ansheng You, Shuai Wang, Fan Wu, and Yang Zheng},
  pdftitle={Mechanism-Aligned Observability for LLM Training Instability}
}

\newcommand{\Wq}{W_q}
\newcommand{\Wk}{W_k}
\newcommand{\dWq}{\Delta W_q}
\newcommand{\dWk}{\Delta W_k}
\newcommand{\Done}{\Delta F}
\newcommand{\Dtwo}{\Delta F_{\text{lin}}}
\newcommand{\Dthree}{R}
\newcommand{\dk}{d_k}

\title{Mechanism-Aligned Observability for Training Instability of Large Language Models}

\author{%
Ruixuan Huang\textsuperscript{1} \quad
Yipei Wang\textsuperscript{2} \quad
Wenyi Fang\textsuperscript{2} \quad
Hantao Huang\textsuperscript{3}\\
\textbf{Ansheng You\textsuperscript{3} \quad
Shuai Wang\textsuperscript{1} \quad
Fan Wu\textsuperscript{2} \quad
Yang Zheng\textsuperscript{2}}\\[0.5em]
\textnormal{\textsuperscript{1}HKUST \quad
\textsuperscript{2}Huawei Technologies \quad
\textsuperscript{3}Independent Researchers}
}

\iclrfinalcopy

\begin{document}

\maketitle
\lhead{Preprint}

\begin{abstract}
Frontier large language model training consumes massive accelerator fleets over long wall-clock times, making stability failures extremely costly. However, global symptoms such as loss divergences often emerge only after the underlying fault has already affected the training dynamics, further increasing the sunk compute. In which internal representation can a training fault become persistently and interpretably observable before it appears in global training metrics? We propose a mechanism-aligned principle for training observability: identify the internal state into which a fault is persistently written, derive the structure induced by its mechanism, and construct monitors in representations aligned with the computation of the affected module. For FlashAttention, we show that biased backward errors accumulate into a persistent low-dimensional structure in weight updates, producing spectral concentration. For MoE routers, we characterize routing dynamics through per-token routing behavior rather than parameter similarity, which may change without altering the routing function. Experiments show that the two monitors respond selectively to changes in their corresponding modules, before degradation becomes visible in global training metrics.
\end{abstract}

\section{Introduction}

Frontier large language model (LLM) training typically occupies thousands of accelerators for weeks to months~\citep{chowdhery2022palm,smith2022megatron}. The parameter counts now reach hundreds of billions to more than a trillion~\citep{yang2025qwen3,kimi2025k2,deepseek2026v4}; pre-training corpora span tens of trillions of tokens~\citep{glm45team2025glm45,longcat2025flash,qwen2026qwen35omni}. DeepSeek-V3 gives an explicit cost accounting, where 14.8T pre-training tokens required 2.788M H800 GPU-hours and a reported \$5.576M in direct rental cost, excluding prior research and ablation experiments~\citep{deepseek2024v3}.

At this scale, training stability has become a first-class engineering concern. Current monitors of training stability still largely rely on global symptoms such as loss collapse. GLM-130B reports loss spikes and divergence in 100B-scale training~\citep{zeng2022glm}; and Kimi K2 introduces MuonClip to address training instability and reports 15.5T-token pre-training with zero loss spike~\citep{kimi2025k2}.
However, the loss collapse is not where a training fault begins, but the downstream outcome of the fault propagating until the system becomes unstable. Before loss collapse appears, a training run may already have entered an unstable state and continue silently. A useful monitoring window is therefore the stage in which a fault has begun to accumulate but has not manifested as loss collapse.

Existing training systems have begun to investigate this window, commonly monitoring gradient norms and parameter norms~\citep{fan2024hlat}. Frontier training also uses attention entropy, maximum attention logit, and router similarity to further characterize training instability~\citep{zhai2023stabilizing,takase2025spike,golden2024flash,kimi2025k2}. 
However, these indicators are expensive to obtain or difficult to relate reliably to the underlying fault. Max-logit signals are difficult to expose in production FlashAttention (FA, \citealp{golden2024flash}) because they require kernel modification or recomputation~\citep{kimi2025k2}.
The same fault can also have different visibility across representations. Consider low-precision FA. Abnormal updates may already be written persistently into the Q/K parameters while the raw weights $W$ remain dominated by initialization. Observing $\Delta W$ can remove this background. But FA does not use Q/K weights independently; it computes through the QK bilinear operator. Examining how the operator changes yields a representation aligned with the fault mechanism. In MoE models, cosine similarity of router weights can reveal whether parameters are becoming redundant, but healthy models can operate at different similarity levels, whereas per-token expert scores and routing distributions directly describe the function executed by routers.

This motivates a mechanism-aligned principle for monitor design
in training observability. We follow the fault-propagation path
to identify the internal state into which the fault first writes
persistent effects; derive the structure that this accumulation
should create; and observe that structure in a representation
aligned with the affected module's behavior.
We instantiate it in two scenarios.

\begin{itemize}[
    leftmargin=0.8em,
    itemindent=0pt,
    topsep=3pt,
    itemsep=3pt,
    parsep=0pt
]
\item \textbf{For low-precision FA,}
we propose update monitoring because low-precision backward
errors first enter the model state through parameter updates.
We decompose the QK operator update and monitor its singular
spectrum.

\item \textbf{For MoE routers,}
the intended computation is discriminative, non-collapsed expert
selection. We therefore monitor per-token routing entropy.
Because it reads the full softmax distribution, it can capture
routing concentration before downstream discrete quantities change.
We further analyze how learning rate, global batch size and
auxiliary loss coefficient interact through stable-winner
reinforcement and accelerate routing concentration.
\end{itemize}

Our experiments show that the proposed monitors
expose structural changes before global training
indicators show degradation.
In one FA fault-injection run, update spectra already depart
from baseline $\sim\!$4,000 steps before the loss upturn,
while loss and gradient norm are still decreasing.
Router per-token entropy reveals routing concentration
under changes in training hyperparameters.
The different responses of the attention and router monitors
provide evidence for attributing observed changes to their mechanisms, supporting early warning and diagnosis of training instability.

\section{Motivation}

Modern training systems can already monitor many internal states, yet existing monitors can still fail to accurately reflect an ongoing fault because of two forms of mismatch: (1) fault-related changes may already be written into parameters while being obscured by the historical background; (2) a substantial change in a parameter-side metric may not correspond to a change in its actual module function. We illustrate them using low-precision FlashAttention and MoE routers to show that the observability of training instability depends on the representation being monitored and on whether that representation is aligned with the underlying fault mechanism and the module’s function.

\subsection{FlashAttention: Fault Signals Masked by Raw Weights}
\label{sec:motivation-fa}

FA fuses the softmax-scaled $QK^\top$ computation
in on-chip memory and does not materialize the full $N \times N$ logit matrix. This brings
dramatic memory and throughput gains, and modern LLM training and inference now rely on
it as the prevalent attention implementation. However, FA is reported as a source of training instability, where low-precision arithmetic in its backward pass
can deposit persistent, biased errors into weight updates~\citep{kimi2025k2, qiu2026why}.
Moreover, the fused implementation blocks the most natural symptom-level monitor used at scale,
namely tracking the maximum attention logit (max-logit) to catch softmax explosion~\citep{kimi2025k2}.
Reading max-logits out of a production FA kernel requires either invasive kernel
modification or a recompute pass, both unacceptable in a large training run.

Runtime training monitors are viable only for quantities that require no kernel modification or
activation recomputation. For example, using weight monitors~\citep{fang2023survey}. However, in practice,
weight-based
indicators are diluted by initialization energy. The natural remaining target is the parameter update $\Delta W$ itself, which
is exactly the level at which low-precision FA faults have been shown to deposit their
persistent damage~\citep{qiu2026why}.

Modern LLMs are severely over-parameterized, with intrinsic dimension far below parameter
count~\citep{jacot2018neural,chizat2019lazy,lee2019wide}, so a fault that perturbs along currently
low-impact directions is absorbed silently. Among parameter monitors, the update $\Delta W$ is preferable to the raw weight $W$ on signal-to-noise grounds: singular-value statistics of $W_t$ are diluted by initialization energy (Appendix~\ref{app:init-dominance}),
whereas the increment $\Delta W_{t,\tau}=W_t-W_{t-\tau}$ removes this background and exposes the
update geometry directly.

\subsection{MoE Routers: Different Similarity Operating Points}
\label{sec:motivation-router}

Many recent LLMs above 30B parameters use MoE architectures~\citep{longcat2025flash,kimi2025k2,wang2024auxiliary}.
In a top-$k$ MoE layer~\citep{shazeer2017outrageously}, a lightweight router
selects which experts process each token. Concretely, with the router weight matrix
$W_R = [w_1, \ldots, w_n]$, each token $x$ receives expert scores
$s = W_R^\top x$, a softmax produces a routing distribution, and the top-$k$ entries dictate which
of the $n$ experts are actually used. Although $W_R$ typically holds well below $0.1\%$ of the
parameters, the choices it makes determine which of the remaining $99.9\%$ are
used on any given token. This asymmetry between trivial parameter count and outsized influence makes the router the natural place to look for MoE instability pathologies.

The community has already used router weight similarity $\operatorname{sim}(W_R)$ for practical training monitoring~\citep{longcat2025flash}. However, the parameter geometry of a router does not uniquely determine its expert-selection behavior. A simple counterexample is to add the same common vector $a$ to every expert weight, which adds $a^\top x$ to every expert score. This can increase weight similarity, while leaving both the softmax routing distribution and the top-$k$ expert selection unchanged. Our scan of open-source MoE checkpoints further shows that even well-functioning models can operate at very different values of $\operatorname{sim}(W_R)$. Router columns in the GPT-OSS family are nearly orthogonal, whereas Qwen3.5-35B-A3B reaches $\sim\!0.51$.
{\color{black} Table~\ref{tab:router-sim-openmodels} further shows that markedly different router-weight similarities can coexist with nearly identical normalized routing entropies.}
This motivates us to focus on the relative expert scores that directly determine expert selection.

\begin{table}[htbp]
\centering
\color{black}
\captionsetup{labelfont={color=black},textfont={color=black}}
\setlength{\tabcolsep}{2pt}
\footnotesize
\newcommand{\routerstat}[2]{\shortstack{$#1$\\[-1pt]$\pm#2$}}
\resizebox{\columnwidth}{!}{%
\begin{tabular}{lccccccccc}
\toprule
Model & \shortstack{GPT-OSS\\20B} & \shortstack{GPT-OSS\\120B} & \shortstack{GLM-4.7\\Flash} & \shortstack{Ling-3.0\\tiny} & \shortstack{Qwen3\\30B-A3B} & \shortstack{Qwen3.5\\35B-A3B} & \shortstack{GLM-5.3\\Flash} & \shortstack{DS-V4-Flash\\preview} & \shortstack{DS-V4-Flash\\0731} \\
\midrule
$\operatorname{sim}(W_R)$ & \routerstat{-0.021}{0.007} & \routerstat{0.000}{0.005} & \routerstat{0.269}{0.027} & \routerstat{0.225}{0.044} & \routerstat{0.492}{0.024} & \routerstat{0.505}{0.080} & \routerstat{0.163}{0.026} & \routerstat{0.180}{0.072} & \routerstat{0.179}{0.072} \\
\midrule
$\bar H/\log n$ (\%) & \routerstat{90.0}{3.7} & \routerstat{94.4}{1.3} & \routerstat{79.4}{3.2} & \routerstat{77.8}{2.3} & \routerstat{87.9}{2.4} & \routerstat{94.4}{1.3} & \routerstat{85.7}{3.8} & \routerstat{49.3}{19.4} & \routerstat{49.8}{20.0} \\
\bottomrule
\end{tabular}}
\caption{Router-weight similarity and normalized per-token entropy across open-source MoE checkpoints, reported as mean$\pm$std across router layers. Entropy is first averaged over prefill tokens within each layer; see Figure~\ref{fig:router-entropy-openmodels} for the probe and layer distributions.}
\label{tab:router-sim-openmodels}
\end{table}
\section{Mechanism-Aligned Monitor Design}

We adopt a mechanism-aligned principle for training monitor design. For a specific training fault, we: (1) identify the internal state into which the fault is persistently written; (2) derive the structure that this persistent effect should induce from the underlying fault mechanism; and (3) construct a monitor in a representation aligned with the actual computation or behavior of the affected module. Through this process, each monitor is tied to an explicit fault mechanism and an interpretable structure. We instantiate this principle below for low-precision FA and MoE routers, showing how different failure mechanisms lead to different internal observables.

\subsection{Attention Update Monitoring}
\label{sec:flash}

As discussed in Section~\ref{sec:motivation-fa}, raw weights can obscure recently written fault signals under historical parameter backgrounds. Based on the known low-precision FA failure mechanism, we therefore introduce the windowed parameter update
$\Delta W_{t,\tau}=W_t-W_{t-\tau}$
as the first monitoring representation.
\citet{qiu2026why} shows that
low-precision FA can induce biased scalar errors in
$\operatorname{rowsum}(dO\odot O)$, and that these biased scalars
multiply structurally coherent rank-one update atoms.
We next ask what observable structure this mechanism should leave in attention updates over a monitoring window.

Let $X\in\mathbb{R}^{N\times d}$
be the input entering the query projection,
$K=XW_k$, and $P=\operatorname{softmax}(\frac{QK^\top}{\sqrt{d_k}})$ be the attention
probability matrix. For a token-step sample $j$, let $X_j$ and $(PK)_j$ denote
the corresponding rows. 
Following Qiu--Yao's mechanism, the update-side source can be written as $e_jR_j$, where $e_j$ is the biased scalar
error induced through $\delta_{\rm FA}=\operatorname{rowsum}(dO\odot O)$ and $R_j=X_j^\top(PK)_j\in \mathbb{R}^{d \times \dk}$ is the associated
rank-one update atom, up to the attention scale and sign.
If $M=\mathbb{E}[e_jR_j]$ has rank $r\ll\dk$ 
and the centered fluctuations $e_jR_j-M$ are independent (or martingale-difference) with bounded second moment, then over $n$ samples
\begin{equation}
\sum_{j=1}^n e_jR_j = nM + O_p(\sqrt n).
\label{eq:accumulation-main}
\end{equation}
The coherent low-rank component grows linearly, while zero-mean residuals grow sublinearly.
Once $n\|M\|_2$ dominates the residual, the update becomes concentrated in the low-dimensional subspace characterized by $M$.
This
turns Qiu--Yao's per-step source model into a
prediction that accumulated $\Delta W$ spectra should develop a low-rank
component. Its
proof is in
Appendix~\ref{app:bias-coherence}.

\textbf{Quantifying Spectral States.}
The structural state of $\Delta W$ can be summarized by various
mathematical quantities. Given its singular values $\sigma_1 \geq \cdots \geq \sigma_r$,
the \emph{stable rank} $\mathrm{srank}(\Delta W) =
\frac{\|\Delta W\|_F^2}{\|\Delta W\|_2^2}$~\citep{ipsen2024stable,
roy2007effective} measures the ratio of the squared Frobenius norm to the squared spectral norm,
which is the inverse fraction of squared spectral energy carried by the largest singular value. However, it cannot distinguish how the remaining energy is distributed across the spectrum.
On the other hand, effective rank $\mathcal{S}_\alpha(\Delta W) = \exp\bigl(-\sum_i p_i \log p_i\bigr)$ with $p_i = \frac{\sigma_i^\alpha}{\sum_j \sigma_j^\alpha}$
is another way of evaluating the state of the spectrum. Empirically, to balance sensitivity and noise, we use
$\alpha=2$, which is also known as \emph{singular spectrum}~\citep{alter2000singular}.

\textbf{The Bilinear Decomposition.}
The $\Delta W$ monitor above treats $\Wq$ and $\Wk$
as independent matrices and computes spectral metrics on $\dWq$,
$\dWk$, or their concatenation.
However, in FA, the attention score
$QK^\top = X(\Wq\Wk^\top)X^\top$ depends on $\Wq$ and $\Wk$ only
through the \emph{bilinear form}
$F(\Wq,\Wk) = \Wq\Wk^\top$,
so monitoring the factors separately can miss correlated drifts that
cancel or amplify in the product.
A natural alternative is to track 
$\Done = F_t - F_{t-\tau}$ that can be decomposed into components with distinct physical
and spectral signatures.
For $W_{q,t}, W_{k,t}$ at two time points and $\dWq = W_{q,t} - W_{q,t-\tau}$,
$\dWk = W_{k,t} - W_{k,t-\tau}$, with the un-subscripted factors evaluated at the base point $t-\tau$,
\begin{equation}
\Done := W_{q,t}W_{k,t}^\top - W_{q,t-\tau}W_{k,t-\tau}^\top
        = \Dtwo + \Dthree,
\label{eq:decomp}
\end{equation}
where $\Dtwo = \dWq \Wk^\top + \Wq \dWk^\top$ and
$\Dthree = \dWq \dWk^\top$.
This follows immediately from the bilinearity of $F$. In particular, $D F[(\dWq, \dWk)] = \Dtwo$ and $\tfrac{1}{2}D^2 F[(\dWq, \dWk)^{\otimes 2}] = \Dthree$, with all higher derivatives vanishing identically.

\textbf{Exact Spectral Computation.}
Although $\Done$, $\Dtwo$, and $\Dthree$ are formally $d\times d$
QK-product increments, their nonzero singular spectra can be computed
exactly from small cores. For any $A,B\in\mathbb{R}^{d\times r}$
with thin QR decompositions $A=Q_A R_A$ and $B=Q_B R_B$,
\begin{equation}
AB^\top = Q_A(R_A R_B^\top)Q_B^\top,
\end{equation}
so the nonzero singular values of $AB^\top$ are those of the
$r\times r$ core $R_A R_B^\top$. Applied to a single attention head,
\begin{equation}
\begin{aligned}
\Dthree &= \dWq\dWk^\top,\\
\Dtwo &= [\dWq,\Wq][\Wk,\dWk]^\top,\\
\Done &= [W_{q,t},W_{q,t-\tau}][W_{k,t},-W_{k,t-\tau}]^\top,
\end{aligned}
\end{equation}
with ranks at most $\dk$, $2\dk$, and $2\dk$, respectively.
Thus monitoring does not require materializing a dense $d\times d$
product when $d\gg\dk$; it only requires the singular spectrum of
a head-dimensional core. Appendix~\ref{app:monitoring-cost} reports the timing
and numerical accuracy of this computation.

The functional increment $\Done$ is the direct monitoring quantity.
Its decomposition provides a mechanism interpretation: $\Dtwo$ maps Q/K
parameter updates into the QK computation to first order, while $\Dthree$
captures their second-order interaction. 
This gives three testable predictions for our evaluation: (1)
the fault-induced spectral concentration should be observable in the windowed
updates; (2) expressing these updates through the QK computation should make the
structure more directly visible than treating $W_q$ and $W_k$ independently;
and (3) when $R$ is small, $\Delta F_{\mathrm{lin}}$ should
closely track the spectral structure of the full QK change. We evaluate these
predictions in Section~\ref{sec:experiments}.

\subsection{MoE Router Monitoring}
\label{sec:router}

As discussed in Section~\ref{sec:motivation-router}, \emph{routing dynamics} determine how the majority of parameters in MoE models are activated, making them a natural functional target for monitoring. Prior work shows that MoE training is particularly sensitive to optimization hyperparameters such as the learning rate (LR) and global batch size (GBS)~\citep{zadouri2023pushing,tian2026hyperparameter,vankadara2026scale}. We therefore first identify the functional quantity that governs routing dynamics, and then derive how its evolution responds to perturbations in LR and GBS. Detailed derivations and their assumptions are provided in Appendix~\ref{app:router-dynamics}.

In matrix form, let
$C_n=I_n-\frac{1}{n}\mathbf{1}\mathbf{1}^\top$ and
$W_{R,c}=W_RC_n=[w_1-\bar w,\ldots,w_n-\bar w]$. Only this centered component
changes the centered logits $\delta(x)=W_{R,c}^\top x$; the common-mode component
adds the same scalar to every expert score and is removed by softmax. Token dispatch is therefore controlled by centered margins
\begin{equation}
    m_{ij}(x)=(w_i-w_j)^\top x.
\end{equation}
Unlike router weight similarity, these margins directly characterize the relative advantage of one expert over another, while removing the common offset ignored by softmax.

\textbf{LR Amplifies Margin Feedback.}
Consider the training objective
$\mathcal L=\mathcal L^{\mathrm{LM}}+\lambda\mathcal R$,
where $\lambda\geq 0$ is the auxiliary loss coefficient (AUX) that controls load balancing.
For fixed input representations and winner--competitor identities,
one plain-SGD step has conditional mean
\begin{equation}
\label{eq:lr}
\mathbb E_t[\Delta m_{j^\star h}]
=
\eta\gamma_h(t)
=
\eta\bigl[
\gamma_h^{\mathrm{LM}}(t)-\lambda\beta_h(t)
\bigr],
\end{equation}
where $\gamma_h^{\mathrm{LM}}$ is task-induced margin feedback
and $\beta_h$ is auxiliary feedback.
Positive $\beta_h$ opposes the current winner.
At a given training state, LR scales the net feedback;
a larger step amplifies margin expansion when $\gamma_h>0$
and margin contraction when $\gamma_h<0$.

Larger updates may promote routing concentration when the net
feedback reinforces the current winner, while a subsequent
shift toward margin contraction may permit entropy recovery.
Appendix~\ref{app:router-transient} provides a qualitative
illustration of these dynamics.

\textbf{GBS Controls Margin Fluctuation.} GBS also affects margin dynamics, but through a different mechanism from LR.
To make the LR--GBS coupling explicit, for global batch size $B$ write the empirical reinforcement
coefficient as
\begin{align}
\widehat{\gamma}_{h,B}(t)=\gamma_h(t)+\xi_{h,B}(t),\qquad
\mathbb{E}[\xi_{h,B}(t)]=0,\qquad
\operatorname{Var}[\xi_{h,B}(t)]\approx \frac{\sigma_h^2(t)}{B}.
\end{align}
Then the stochastic component of the one-step margin update obeys
\begin{equation}
    \operatorname{Var}[\Delta m_{j^\star h}(t)]
\approx \eta^2\operatorname{Var}[\xi_{h,B}(t)]
\approx \frac{\eta^2\sigma_h^2(t)}{B}.
\end{equation}
Thus a smaller GBS increases the per-step margin variance at fixed LR.
If the trajectory
additionally enters the positive-$\gamma_h(t)$ stable-winner regime above, these noise-induced margin excursions
may be reinforced over subsequent steps.

{\color{black}
\textbf{Quantifying Routing Concentration.}
The preceding analysis characterizes conditional margin drift and
sampling fluctuations under the stated assumptions. These local
relations do not by themselves establish monotone concentration
or an ordering across training trajectories.
To observe how the routing distributions evolve during training,
we monitor the per-token entropy of the full softmax distribution
before top-$k$ dispatch:
\begin{equation}
H(p(x))=-\sum_{i=1}^{n}p_i(x)\log p_i(x),
\qquad
p(x)=\operatorname{softmax}(W_{R,c}^{\top}x).
\label{eq:router-entropy}
\end{equation}
Entropy provides a continuous summary of probability concentration.
It complements discrete assignment and aggregate load statistics,
such as MaxVio~\citep{wang2024auxiliary}, because the routing
probabilities can become more concentrated while the selected
expert set and aggregate assignment counts remain unchanged.

This observable measures concentration in the pre-dispatch routing distribution. In the top-\(k\) setting studied here, sustained entropy collapse toward zero indicates that the mixture weights become dominated by a single expert, even though multiple experts are dispatched. Softmax saturation also attenuates the local sensitivity of routing probabilities to logit updates. Aggregate load balance does not exclude this state, because different tokens can saturate toward different experts. We therefore monitor sustained entropy collapse as a warning signal for routing saturation during training.
}

\section{Experimental Evaluation}

\label{sec:experiments}

\subsection{Experimental Setup}

\label{sec:experimental-setup}

\textbf{Models.}
We train a Qwen3-like MoE model with
8 billion total and 1 billion activated parameters per token (8B-A1B) with the proposed monitors.
Its 28 layers have hidden size 1,152, and 128 routed
experts with top-6 dispatch. The upper bound of the router per-token entropy is $\log 128\approx4.852$. This is more representative of modern LLM training than
the GPT-2-small setting in prior work.

\textbf{FA Fault Injection.}
\label{sec:fa-fault-injection}
We compare baseline training with mantissa truncation
at the FA backward inputs $O$ and $dO$~\citep{li2018exploration,qiu2026why}.
We denote each truncation setting by FB$b$, where $b$ is the number of
retained explicit fraction bits in both FP32 $O$ and BF16 $dO$.
The three settings are {FB2}, {FB1}, and {FB0};
Appendix~\ref{app:fa-fault-injection} gives the implementation and error expansion.

\textbf{Training Settings.}
For FA experiments, we use LR $5\times10^{-4}$, GBS 384, and auxiliary-loss
coefficient (AUX) 0.001 in both baseline and fault-injected runs. For router experiments, we
perturb training settings: the LR sweep uses
$\{5\times10^{-5},10^{-4},10^{-3},10^{-2},5\times10^{-2}\}$ with AUX
$\{0,0.001\}$ at GBS 384; the GBS sweep uses $\{24,96,384\}$ with
LRs $\{10^{-3},10^{-2}\}$ at AUX 0.001.

\begin{figure}[h]
\centering
\begin{subfigure}[b]{0.98\linewidth}
    \centering
    \includegraphics[width=\linewidth]{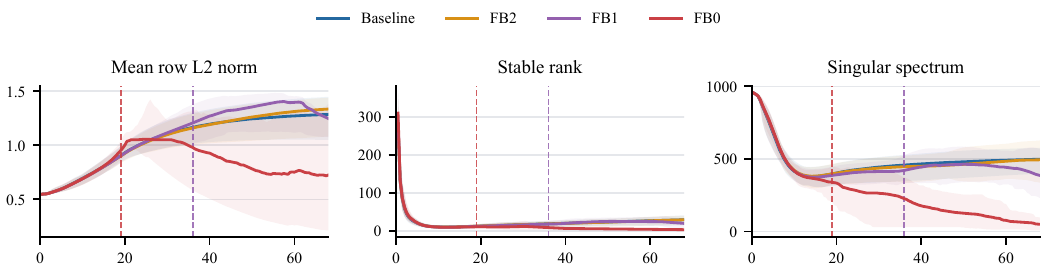}

\caption{QKV weight $W=[W_Q;W_K;W_V]$.}

    \label{fig:W_metrics}
\end{subfigure}
\begin{subfigure}[b]{0.98\linewidth}
    \centering
    \includegraphics[width=\linewidth]{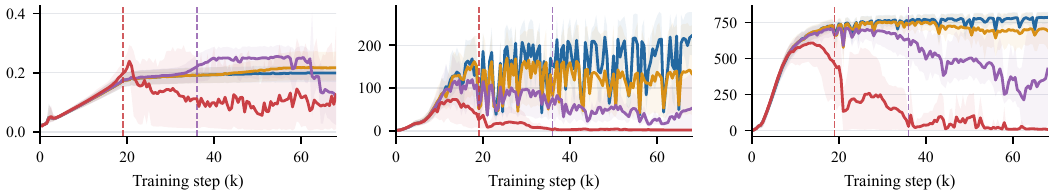}

\caption{QKV weight update $\Delta W=W_t-W_{t-500}$.}

    \label{fig:dW_metrics}
\end{subfigure}
\begin{subfigure}[b]{0.98\linewidth}
    \centering
    \includegraphics[width=\linewidth]{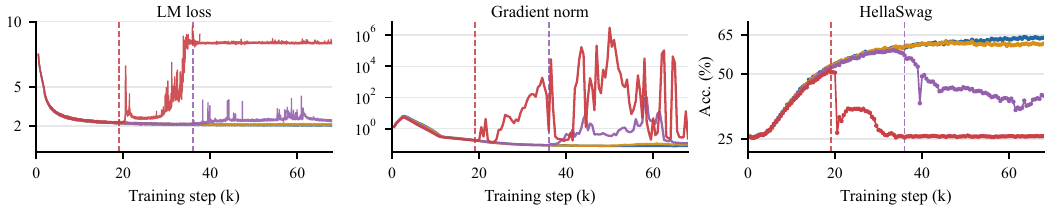}

\caption{Training loss, gradient norm, and checkpoint evaluation.}

    \label{fig:figure1-shared-loss}
\end{subfigure}
\caption{(a,b) Weight statistics for baseline and FA injection
settings. Curves show means across layers. Shaded bands indicate the P10--P90 range across layers.
(c) Loss, smoothed gradient norm, and zero-shot HellaSwag~\citep{zellers2019hellaswag}  evaluation accuracy. The vertical dashed lines mark the FB0 and FB1 loss upturns at steps $
\sim\!$19,000 and 36,000.}

\label{fig:w-dw-metrics}
\end{figure}

\subsection{Experiments on Low-Precision FA}

\label{sec:eval-fa}

The accumulation mechanism predicts that coherent arithmetic errors can
concentrate the spectrum of recent attention updates before substantially
changing the accumulated weights. We evaluate this in three steps:
compare magnitude and spectral summaries of weights and updates, follow the
updates into QK functional space, and align their trajectories with loss,
gradient norm, and checkpoint evaluation. The goal is to determine whether
the predicted structure is visible before overt training degradation and
which representation makes that structure interpretable.

\label{sec:eval-w-vs-dw}

\textbf{Weight vs. Its Update.} Figure~\ref{fig:w-dw-metrics} compares baseline, FB2, FB1, and FB0
using concatenated $W=W_{QKV}=[W_Q;W_K;W_V]$. Its update is
$\Delta W_t=W_t-W_{t-\tau}$, where $\tau=500$. We compare mean row $L_2$ norm, stable rank, and singular spectrum.
The norm measures update magnitude. Stable rank measures total spectral
energy relative to the largest singular value. Singular spectrum
uses the distribution across all singular values, which therefore
matches the predicted redistribution of update energy more directly than a
magnitude-only statistic.

In the baseline run, update spectral diversity
initially grows and later levels off; a sustained reversal relative to that
trajectory indicates concentration. The interpretation depends on deviation
from the baseline trajectory, since healthy
updates can themselves be low rank.
In FB0, the update geometry departs from baseline while the raw weights
remain close to it. At step 15,000, the update stable rank and singular
spectrum are 51\% and 87\% of baseline,
whereas the raw-weight values remain 96\% and 97\%.
The update norm is 104\% of baseline. Monitors on updates remove the accumulated background and expose the recent
structural change.

\label{sec:eval-dw-vs-qk}

\begin{figure}[!htbp]
\centering
\includegraphics[width=0.98\linewidth]{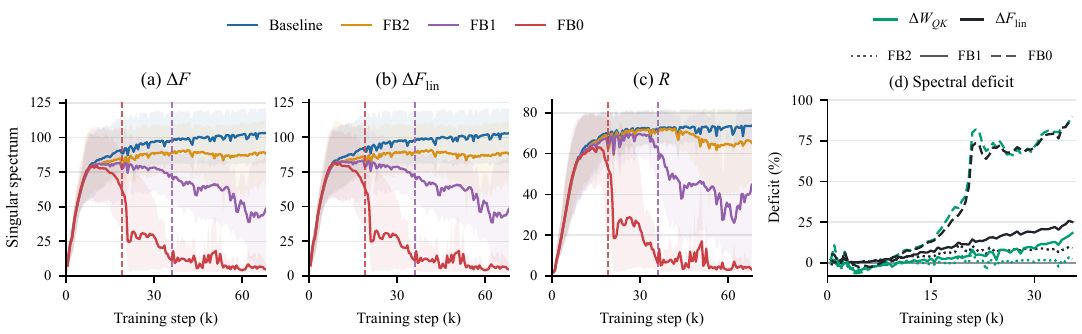}

\caption{QK functional-update spectra and comparison with parameter
updates. (a--c) Singular spectrum of $\Done$, $\Dtwo$, and $\Dthree$,
using the same runs as Figure~\ref{fig:w-dw-metrics}. Curves show means across layers. Shaded bands indicate the P10--P90 range across layers.
(d) Baseline-relative spectral deficit $D$.
}

\label{fig:qk-increment-metrics}
\end{figure}

\textbf{Parameter Updates vs. QK Functional Updates.} 
We next ask whether mapping weight
updates into QK functional space makes the fault structure
more apparent. To isolate the QK
computation, we extract the Q/K blocks and use their concatenated
update $\Delta W_{QK}=[\Delta W_Q;\Delta W_K]$ as the parameter-side
reference. The first-order functional increment
$\Dtwo$
expresses how these updates change the
QK operator. Figure~\ref{fig:qk-increment-metrics}(a--c) shows the
singular spectrum of $\Done$, $\Dtwo$, and $\Dthree$
under baseline and FB0--FB2.
Panel (d) compares the monitoring deviation of $\Dtwo$ and
$\Delta W_{QK}$.

In FB0, parameter and functional
updates exhibit spectral deviations of similar magnitude,
consistent with the fault structure already being visible
in both representations. The benefit of $\Dtwo$ is more pronounced under the milder FB1 and FB2
settings. Figure~\ref{fig:qk-increment-metrics}(d) compares
their baseline-relative spectral deficits
$D=1-S^{\mathrm{FB}}/S^{\mathrm{Base}}$. Here, $S$ denotes the
singular spectrum and a larger positive deficit
indicates stronger spectral concentration. At step 35,000,
the deficit increases from 16.6\% for $\Delta W_{QK}$ to 25.8\%
for $\Dtwo$ under FB1, and from 3.5\% to 9.9\% under FB2.
These results show that QK functional updates provide more
pronounced early-warning signals of training instability.

\label{sec:eval-pre-divergence}

\textbf{Preemptive Observability.}
Figure~\ref{fig:w-dw-metrics}(c) places the spectral deviations
in the context of training progress. Under FB0, the update
spectra have already departed from baseline by step 15,000,
while loss and gradient norm are still decreasing and
checkpoint evaluation accuracy continues to improve.
The structural warning thus precedes the loss upturn near
step 19,000, despite apparently favorable training and
evaluation trends. Under FB1, the functional spectral deviation
is also visible before the
loss upturn at step 36,000. These observations identify
a monitoring window in which structural changes
are detectable before global training metrics show
obvious degradation.

\subsection{Experiments on MoE Routing}

\begin{figure}[!htbp]
\centering
\includegraphics[width=0.98\linewidth]{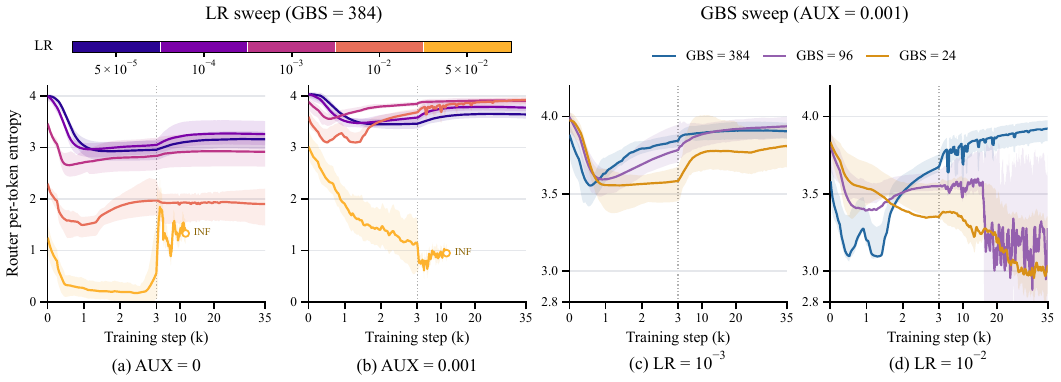}
\caption{
Router per-token entropy under LR and GBS variation.
(a, b) LR sweeps. INF marks the point at which the LM loss becomes infinite and training fails.
(c, d) GBS sweeps.
Curves show means across layers. Shaded bands indicate the P10--P90 range across layers.
Steps 0--3,000 and 3,000--35,000 occupy equal
horizontal widths, each interval with a linear scale.
}
\label{fig:entropy}
\end{figure}

\textbf{LR and Routing Concentration.}
\label{sec:eval-router-lr}
Section~\ref{sec:router} predicts that LR scales the net margin
feedback, combining task-induced reinforcement with auxiliary
balancing. When the net feedback favors the current winner,
larger updates amplify margin separation and concentrate routing
probability. Figure~\ref{fig:entropy}(a) shows the clearest
correspondence without auxiliary balancing: across the tested
LRs, higher-LR runs exhibit an earlier and deeper entropy decline
and retain lower entropy throughout subsequent training.
With AUX enabled, Figure~\ref{fig:entropy}(b)
shows the LR-dependent ordering during
the first 1,000 steps. Then, several runs exhibit entropy
recovery, consistent with auxiliary feedback counteracting
task-induced reinforcement as training evolves.
However, at LR $5\times10^{-2}$, router per-token entropy
{\color{black} remains low after its initial decline,}
showing persistent routing concentration despite auxiliary
balancing. This provides early warnings of training instability
and supports that auxiliary balancing can
promote recovery without guaranteeing protection against
excessively large updates.

\textbf{GBS and Margin Fluctuations.}
\label{sec:eval-router-gbs}
Section~\ref{sec:router} predicts that smaller batches increase
the stochastic variance of margin updates at a fixed
instantaneous LR. When these margin excursions reinforce
the current winner, they can promote persistent routing
concentration. Figure~\ref{fig:entropy}(c) shows that, at LR $10^{-3}$, GBS 24 exhibits a more persistent
entropy reduction, while GBS 96 and 384 recover to similar,
higher-entropy levels.
The contrast is clearer at LR
$10^{-2}$. As shown in Figure~\ref{fig:entropy}(d), GBS 384 recovers after its initial entropy decline,
whereas GBS 96 and 24 develop substantially lower entropy
later in training. These observations support the predicted
susceptibility to routing concentration under smaller batches
and larger updates, with per-token entropy exposing concentration.

\subsection{Selective Signatures Across Modules}

\label{sec:eval-selectivity}

A practical question for training-time monitoring is whether
these responses help identify the mechanisms underlying an
observed instability.
This requires module-specific selectivity:
each observable should preferentially reflect changes in the
function it is designed to monitor. Such selectivity would
strengthen the connection between the observable and its
intended functional mechanism, while providing evidence for
attributing an observed instability to its likely source.
We therefore compare FA- and router-side responses
under FA fault injection and changes in router auxiliary balancing.

\label{sec:eval-selectivity-fa}

\begin{figure}[!htbp]
\centering
\includegraphics[width=0.98\linewidth]{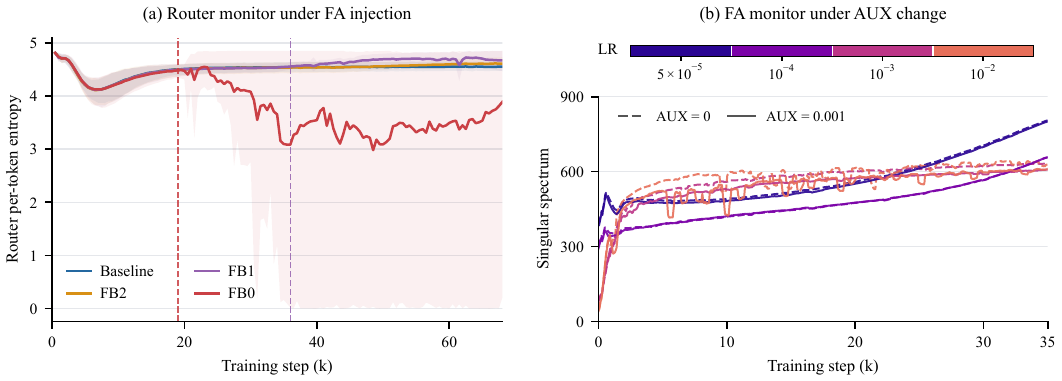}
\caption{Cross-module responses to FA and AUX perturbations.
(a) Router per-token entropy under FA fault injection; colors
follow Figure~\ref{fig:w-dw-metrics}; shaded bands indicate the P10–P90 range across layers. (b) FA singular spectrum $\Delta F_{\text{lin}}$ at GBS 384. Curves show means across layers.}
\label{fig:signature}
\end{figure}

The attention-update spectra separate from baseline
while router entropy initially remains close to it. 
At step 15,000, FB0's router entropy differs
from baseline by less than 0.6\% (Figure~\ref{fig:signature}(a)), whereas its singular
spectrum $\Delta F_{\mathrm{lin}}$ is $\sim\!$13\% lower (Figure~\ref{fig:qk-increment-metrics}(b)). Conversely, changing AUX produces a substantially larger
relative response in router entropy than in the FA monitors, see Figures~\ref{fig:entropy}(a,b) and~\ref{fig:signature}(b).
At LR $5\times10^{-5}$ and $10^{-4}$, the FA curves for AUX 0
and AUX 0.001 nearly coincide, with NRMSEs of only 1.73\%
and 1.56\%, respectively, whereas the corresponding router
entropy NRMSEs are 15.78\% and 16.13\%.
These results show that each monitor responds more strongly to perturbations of its target module, which helps localize the source of an observed change.

\section{Related Work}

\paragraph{Attention Monitoring.}
Prior work diagnoses and stabilizes attention through its logits,
entropy, and numerical behavior~\citep{dehghani2023scaling,wortsman2023smallscale}.
Kimi K2 uses QK-clipping through MuonClip to prevent excessive
attention logits~\citep{kimi2025k2}.
\citet{zhai2023stabilizing} connect attention-entropy collapse to training
instability and propose $\sigma$Reparam based on spectral normalization.
\citet{takase2025spike} analyze loss spikes through the spectral norms of
sub-layer Jacobians.
For FlashAttention, \citet{golden2024flash} quantify numerical deviations
and investigate their downstream effects, while \citet{qiu2026why}
identify a failure mechanism involving biased rounding errors and
similar low-rank representations that corrupt weight updates.
\citet{bao2024selfattention,pan2024dissecting} relate QK geometry to attention localization
and singular-vector structure.

\paragraph{Router Monitoring.}
MoE training commonly uses auxiliary objectives to balance expert
loads~\citep{shazeer2017outrageously,fedus2021switch}.
\citet{wang2024auxiliary} analyze interference from balancing gradients
and propose loss-free balancing through dynamically adjusted expert
biases. For router diagnosis, LongCat-Flash monitors pairwise cosine
similarity of expert router weights and the relative gradient norms
of balancing and language-modeling objectives on batch-averaged expert
probabilities~\citep{longcat2025flash}.
Behavior-side analyses also identify routing uncertainty. GW-MoE
studies tokens with nearly equal expert scores and mitigates their
sensitivity to expert selection~\citep{wu2024gwmoe}.

\section{Conclusion}

We derived internal training monitors for two stability-critical modules
of modern LLMs by asking what each module is supposed to compute and
where damage from its known failure mechanism would first appear.
For FA, biased low-precision backward errors produce coherent low-rank drift
in accumulated weight updates. Monitoring the spectrum of the
exact QK functional increment $\Done$ makes this structure more directly
visible. For MoE routers, the answer is per-token entropy.
Modern LLM architectures continue to introduce modules with their own internal dynamics. As these modules enter production training, each will need monitors derived from its own mechanism. The methodology demonstrated here provides a template for that derivation.

\bibliographystyle{iclr2027_conference}
\bibliography{refs}

\newpage

\appendix
\section*{Appendix}

\section{Formulas}
Table~\ref{tab:observable-formulas} summarizes computational formulas not explicitly provided in the main text.

\begingroup
\captionsetup{hypcap=false}
\captionof{table}{Computation of the main observables and supporting quantities.}
\label{tab:observable-formulas}
\setlength{\tabcolsep}{5pt}
\renewcommand{\arraystretch}{1.28}
\noindent
\begin{tabular}{@{}p{0.25\linewidth}p{\dimexpr0.75\linewidth-10pt\relax}@{}}
\toprule
Quantity & Definition or computation \\
\midrule
FA backward quantities &
$dO=\partial\mathcal L/\partial O,\quad dP=\partial\mathcal L/\partial P=dOV^\top$;\newline
$\delta_{\rm FA}=\operatorname{rowsum}(dO\odot O)$,
one scalar per token. \\
Mean row $L_2$ norm &
$\displaystyle N(A)=\frac1m\sum_{j=1}^{m}\|A_{j,:}\|_2$. \\
Router weight similarity &
$\displaystyle \operatorname{sim}(W_R)
=\frac{1}{n(n-1)}\sum_{i\ne j}
\frac{w_i^\top w_j}{\|w_i\|_2\|w_j\|_2}$,
for nonzero $w_i$. \\
NRMSE &
$\displaystyle \operatorname{NRMSE}(a,b)
=\sqrt{\frac{\sum_{t}(b_t-a_t)^2}
{\sum_{t}a_t^2}}$.\\
\bottomrule
\end{tabular}
\par
\endgroup

\section{Initialization Dominance and Raw-Weight Monitoring}
\label{app:init-dominance}

We justify the initialization-dominance claim used in Section~\ref{sec:flash}. Let $W_t=W_0+E_t$, where $E_t$ is the total displacement from initialization. Weyl's singular-value perturbation inequality gives, for every singular index $i$,
\begin{equation}
|\sigma_i(W_t)-\sigma_i(W_0)| \leq \|E_t\|_2 \leq \|E_t\|_F.
\end{equation}
Thus, when $\|E_t\|_F \leq \varepsilon \|W_0\|_F$ with $\varepsilon<1$, every singular-value statistic of the raw weight is observed through an initialization-dominated background. This does not say that the network is not learning: in lazy or NTK-like regimes, function values can change while parameter displacement remains small. It only says that a raw-weight monitor has poor signal-to-noise for faults that first alter the update geometry. The update increment $\Delta W_{t,\tau}=W_t-W_{t-\tau}$ removes $W_0$ exactly, so spectral concentration in the update is not diluted by initialization energy.

\section{Accumulation Consequence for FA}
\label{app:bias-coherence}

This section proves the concentration form of the accumulation consequence
(Equation~\ref{eq:accumulation-main}) used for the monitoring window. The source model follows the per-step mechanism of~\citet{qiu2026why}: low-precision FA supplies biased scalar
coefficients $e_j$ multiplying coherent rank-one update atoms
$R_j=X_j^\top(PK)_j$.

Index token-step samples by $j$ and set
$R_j=X_j^\top(PK)_j \in \mathbb{R}^{d \times \dk}$. Assume
$Y_j=e_jR_j-M$ are independent or martingale-difference fluctuations with
$M=\mathbb{E}[e_jR_j]$ and $\mathbb{E}\|Y_j\|_F^2\leq \nu^2$. Then
\begin{equation}
A_n := \sum_{j=1}^n e_jR_j = nM + Z_n, \qquad Z_n:=\sum_{j=1}^n Y_j .
\end{equation}
By orthogonality of the centered increments,
\begin{equation}
\mathbb{E}\|Z_n\|_F^2 \leq n\nu^2.
\end{equation}
Markov's inequality implies $\|Z_n\|_F=O_p(\sqrt n)$, hence $\|Z_n\|_2=O_p(\sqrt n)$. This proves Equation~\ref{eq:accumulation-main}. The biased mean $nM$ grows linearly, while the zero-mean residual grows sublinearly.

Finally, suppose $M$ has rank $r$ and singular gap $\sigma_r(M)>0$. Weyl's inequality gives
\begin{equation}
|\sigma_i(A_n)-n\sigma_i(M)| \leq \|Z_n\|_2 .
\end{equation}
When $n\sigma_r(M)\gg \|Z_n\|_2$, the top $r$ singular values of $A_n$ are controlled by $M$ and the remaining singular values are residual-scale. Consequently, stable rank and singular spectrum of the accumulated update converge toward those of the low-rank mean $M$. This is the formal sense in which biased low-precision arithmetic becomes a low-rank $\Delta W$ fault under coherence of rank-one atoms.

\paragraph{FA fault injection.}

\label{app:fa-fault-injection}

The injection clears low-order fraction bits by reinterpreting each
selected floating-point value as an integer storage word $u$ of the same
width, applying a bit mask, and reinterpreting the result in the original
dtype. For a mask width $n$, the stored-bit operation is
\begin{equation}
\tilde u = u\mathbin{\&}\operatorname{mask}(n),\qquad
\operatorname{mask}(n)=\underbrace{11\cdots11}_{\text{retained bits}}\underbrace{00\cdots00}_{n\text{ bits}}.
\end{equation}

In our training runs, $O$ is retained in FP32 and $dO$ in BF16 at the
injection sites. FB$b$, for $b\in\{2,1,0\}$, uses $n_O=23-b$ and
$n_{dO}=7-b$, leaving $b$ explicit fraction bits in both tensors while
preserving sign and exponent. FB0 removes all explicit fraction bits;
it does not remove the implicit leading bit of normalized values.
FB2, FB1, and FB0 correspond to the original log names O21/dO5,
O22/dO6, and O23/dO7, respectively. Clearing fraction bits truncates finite values
toward zero and creates deterministic, input-dependent error; it does not
implement round-to-nearest. Reduced-precision truncation is a numerical
emulation method used by \citet{li2018exploration}. The selection of backward
FA tensors is motivated by \citet{qiu2026why}, but the exact masks here are
our controlled stress test, not a protocol established by that work.
For a BF16 example with $n=3$,
\begin{equation}
\begin{array}{rcl}
\text{original value } 1.1015625
&:&
\text{\texttt{0|01111111|0001101}}_{\mathrm{BF16}}
=\text{\texttt{0x3f8d}}
\\[2pt]
&& \downarrow\ \text{\texttt{(uint16 >> 3) << 3}}
\\[2pt]
\text{truncated value } 1.0625
&:&
\text{\texttt{0|01111111|0001000}}_{\mathrm{BF16}}
=\text{\texttt{0x3f88}} .
\end{array}
\end{equation}
The exact tensor and mask width are experimental knobs, but the resulting
tensor-level error has the same algebraic role. If the backward path uses
truncated tensors $\widehat O=O+E_O$ and
$\widehat{dO}=dO+E_{dO}$, the scalar source can include perturbations to both
$O$ and $dO$:
\begin{equation}
\widehat\delta_{\text{FA}}-\delta_{\text{FA}}
=\operatorname{rowsum}(dO\odot E_O+E_{dO}\odot O+E_{dO}\odot E_O).
\end{equation}
This implementation-specific expansion only changes what contributes to the
scalar $e_j$. Perturbing $dO$ also induces a direct perturbation of
$dP=dOV^\top$, so the main text uses only the abstract source form $e_jR_j$
rather than treating the $O,dO$ expansion as a separate mechanism.

\subsection{Magnitude Regime of QK Functional Increments}

\label{app:qk-magnitude-regime}

With the un-subscripted factors evaluated at $t-\tau$, bilinearity gives
\begin{equation}
\begin{aligned}
\Done
&=(\Wq+\dWq)(\Wk+\dWk)^\top-\Wq\Wk^\top\\
&=\underbrace{\dWq\Wk^\top+\Wq\dWk^\top}_{\Dtwo}
 +\underbrace{\dWq\dWk^\top}_{\Dthree}.
\end{aligned}
\end{equation}

Let $w=\max(\|\Wq\|_F,\|\Wk\|_F)$ and
$u=\max(\|\dWq\|_F,\|\dWk\|_F)$. Submultiplicativity and the triangle
inequality imply
\begin{equation}
\begin{aligned}
\|\Dtwo\|_F
&\leq \|\dWq\|_F\|\Wk\|_2+\|\Wq\|_2\|\dWk\|_F
\leq 2wu,\\
\|\Dthree\|_F
&\leq \|\dWq\|_F\|\dWk\|_2\leq u^2.
\end{aligned}
\end{equation}

Thus the first-order and second-order terms have upper bounds of order $wu$
and $u^2$, respectively. Upper bounds alone do not establish dominance: the
two first-order contributions can cancel, or their products can be small.
To make the magnitude regime explicit, assume
$\|\Dtwo\|_F\geq cwu$ for a constant $c>0$ independent of $u/w$.
This condition rules out near cancellation or degenerate first-order action.
Then
\begin{equation}
\frac{\|\Done-\Dtwo\|_F}{\|\Dtwo\|_F}
=\frac{\|\Dthree\|_F}{\|\Dtwo\|_F}
\leq \frac{u}{cw}.
\end{equation}

In the early-to-mid training regime $u\ll w$, the exact increment is therefore
well approximated by its first-order term under this condition. Moreover,
singular-value perturbation gives
\begin{equation}
|\sigma_i(\Done)-\sigma_i(\Dtwo)|
\leq \|\Dthree\|_2\leq u^2.
\end{equation}

This explains why the exact increment and its linear term can show similar
spectral behavior, without guaranteeing identical values of a normalized
spectral statistic or identical detection times. The second-order interaction
can still have a structured spectrum despite its smaller magnitude; its role
is to characterize Q/K update coupling within the decomposition.
The empirical comparison in Section~\ref{sec:eval-dw-vs-qk} uses $\Done$ as
the direct monitor and $\Dtwo$ and $\Dthree$ to interpret that monitor.

\section{The Router Weight--Entropy Link}
\label{app:entropy}
\label{app:router-dynamics}

This appendix connects router updates to margins and per-token entropy, including auxiliary balancing and batch noise.
We analyze plain SGD with LR $\eta>0$ on a fixed probe distribution.
For $n\ge2$ experts, let $W_R=[w_1,\ldots,w_n]\in\mathbb R^{d\times n}$, $C_n=I_n-\mathbf{1}\mathbf{1}^\top/n$, and $W_{R,c}=W_RC_n$.
The logits are $s_i(x)=w_i^\top x$, their centered vector is $\delta(x)=W_{R,c}^\top x$, and $p(x)=\operatorname{softmax}(\delta(x))$.
Centering leaves softmax unchanged.
We use $H(p)=-\sum_i p_i\log p_i$ with natural logarithms, and write $\mathbb E_t$ for expectation conditional on the state before sampling a minibatch.

\paragraph{Local weight--entropy relation.}
For centered logits near zero,
{
\begin{equation}
H(\operatorname{softmax}(\delta))
=\log n-\frac{\|\delta\|_2^2}{2n}+R_H(\delta),
\qquad R_H(\delta)=O(\|\delta\|_2^3).
\label{eq:entropy-expansion}
\end{equation}
}
Writing $H=\log Z-p^\top\delta$ with $Z=\sum_i e^{\delta_i}$, Taylor expansion at centered $\delta=0$ gives zero gradient and Hessian $-C_n/n$, yielding Equation~\ref{eq:entropy-expansion}.
For fixed probe second moment $M_x=\mathbb E_x[xx^\top]$, taking expectations gives
\begin{equation}
\log n-\mathbb E_x H(p(x))
=\frac1{2n}\operatorname{tr}(W_{R,c}^\top M_xW_{R,c})
+O(\mathbb E_x\|\delta(x)\|_2^3).
\label{eq:weight-entropy-link}
\end{equation}
This assumes probe logits in the local neighborhood and finite displayed moments.
For $m_{ij}=s_i-s_j$, centering gives $\sum_{i<j}m_{ij}^2=n\|\delta\|_2^2$; hence the quadratic term also equals $\sum_{i<j}\mathbb E_xm_{ij}^2/(2n^2)$.
This connects entropy to the energy of all margins without assuming isotropy.
The expansion cannot be extrapolated to saturation: the true entropy deficit is bounded by $\log n$, whereas the quadratic term can grow without bound.

\subsection{How a Larger LR Can Reinforce the Winner}
\label{app:router-feedback}

This section derives the mechanism behind \emph{LR Amplifies Margin Feedback} (Section~\ref{sec:router}).
By \emph{winner reinforcement}, we mean an increase in the logit gap between the currently highest-scoring expert and a competitor.
A larger LR amplifies this increase when the combined task and auxiliary feedback favors the winner at a fixed training state.
We first derive these two feedback contributions, then connect their net effect to the winner's routing probability.

\paragraph{Task and auxiliary contributions to winner reinforcement.}
Let $\mathcal L^{\mathrm{LM}}_{\mathcal B}=M^{-1}\sum_{u\in\mathcal B}\ell_u$ be the LM loss over $M$ tokens and $\mathcal L_{\mathcal B}=\mathcal L^{\mathrm{LM}}_{\mathcal B}+\lambda\mathcal R_{\mathcal B}$, with $\lambda\ge0$.
For the dense-softmax relaxation $y(x)=\sum_i p_i(x)E_i(x)$~\citep{jacobs1991adaptive,shazeer2017outrageously}, with expert outputs $E_i(x)$, define $q_i=M\langle\partial\mathcal L^{\mathrm{LM}}_{\mathcal B}/\partial y,E_i(x)\rangle$ and $\bar q=\sum_i p_iq_i$.
Holding expert outputs fixed, the chain rule gives $r_i^{\mathrm{LM}}:=M\,\partial\mathcal L^{\mathrm{LM}}_{\mathcal B}/\partial s_i=p_i(q_i-\bar q)$.
Thus $q_i$ is a local loss sensitivity, and a current winner need not receive reinforcing task feedback.

For the auxiliary-specific results, assume the stop-gradient load-balancing objective
\begin{equation}
\mathcal R_{\mathcal B}=n\sum_i\operatorname{sg}(f_i)P_i,
\qquad P_i=\frac1M\sum_{x\in\mathcal B}p_i(x),
\qquad f_i=\frac1{Mk}\sum_{x\in\mathcal B}\mathbf{1}\{i\in\operatorname{TopK}(x)\}.
\label{eq:router-aux-objective}
\end{equation}
Exactly $k$ assignments per token imply $\sum_i f_i=1$ and $f_i\le1/k$.
Loads are recomputed each forward pass and held fixed during differentiation.
Writing $\bar f_p=\sum_i p_if_i$, differentiation gives
\begin{equation}
a_i:=M\frac{\partial\mathcal R_{\mathcal B}}{\partial s_i}
=np_i(f_i-\bar f_p),\qquad
r_i=r_i^{\mathrm{LM}}+\lambda a_i,\qquad
\widehat G=\nabla_{W_R}\mathcal L_{\mathcal B}
=\frac1M\sum_{x\in\mathcal B}x\,r(x)^\top.
\label{eq:router-aux-gradient}
\end{equation}
Both logit-gradient components sum to zero, so $\widehat G C_n=\widehat G$.
For balancing groups $\mathcal B_g$ of size $M_g$, use $\mathcal R=\sum_g(M_g/M)\mathcal R_g$ and each token's group loads in $a_i$; the same $1/M$ normalization then holds.
The balancing group need not equal the global batch.

\paragraph{How LR amplifies the winner's advantage.}
Fix a probe $x'$ and winner--competitor identities $j^\star,h$, with margin $m_{j^\star h}=(w_{j^\star}-w_h)^\top x'$.
One SGD step gives
{
\begin{equation}
\Delta m_{j^\star h}
=-\frac\eta M\sum_{x\in\mathcal B}(x'^\top x)(r_{j^\star}-r_h)
=\eta\widehat\gamma_h,\qquad
\mathbb E_t\Delta m_{j^\star h}
=\eta(\gamma_h^{\mathrm{LM}}-\lambda\beta_h)=:\eta\gamma_h.
\label{eq:router-margin-update}
\end{equation}
}

Here $\gamma_h^{\mathrm{LM}}$ and $\beta_h$ are the conditional means of the corresponding negative task-gradient and positive auxiliary-gradient batch averages, respectively, including $1/M$ but excluding $\eta$ and $\lambda$.
Positive $\beta_h$ opposes the winner, but an underloaded winner can instead receive auxiliary reinforcement.
Thus, when $\gamma_h^{\mathrm{LM}}>\lambda\beta_h$, increasing LR at the same state produces a larger conditional mean increase in the winner's margin.
Across LR runs, the net feedback can change sign and evolve differently, so this one-step relation does not order entire trajectories.
For fixed ordering, strongest competitor $h^\star$, and constant LR, write $m_t=m_{j^\star h^\star}(x',t)$; then
{
\begin{equation}
\begin{aligned}
m_T=m_0+\eta\sum_{t<T}\gamma_{h^\star}(t)
+\eta\sum_{t<T}\xi_{h^\star}(t),\\
\xi_h:=\widehat\gamma_h-\gamma_h,\quad\mathbb E_t\xi_h=0,\quad
1-p_{j^\star}(T)&\le(n-1)e^{-m_T}.
\end{aligned}
\label{eq:router-pathwise-concentration}
\end{equation}
}

The bound follows by factoring the winner's exponential out of the softmax denominator.
{\color{black}
Positive conditional mean feedback does not by itself imply
positive realized cumulative margin growth, because the accumulated
sampling fluctuations can be negative.
The recurrence therefore describes realized margin changes without
guaranteeing monotone concentration along stochastic training
trajectories.
}

Cyclic permutations of a probability vector with distinct entries and maximum approaching one give $f_i=P_i=1/n$ and $a=0$, while every token's entropy approaches zero; load balance therefore supplies no per-token entropy floor.

\subsection{Transient Recovery and a Finite-Step Stability Boundary}
\label{app:router-transient}

When loads are balanced, the auxiliary derivative vanishes; subsequent load and task-gradient changes can reverse the net feedback.
To illustrate recovery, assume a probe follows a fixed centered, nonconstant direction $\delta_t=z_tv$, $z_t\ge0$.
Along this direction, $dH/dz=-z\operatorname{Var}_{i\sim p}(v_i)\le0$.
Suppose task reinforcement is $A\rho^t+a_1z_t$ and auxiliary restoration is $\lambda bz_t$, with $A,b>0$, $a_1\ge0$, $0<\rho<1$, and $\kappa:=\lambda b-a_1>0$.
Then
\begin{equation}
z_{t+1}=(1-\eta\kappa)z_t+\eta A\rho^t,
\qquad z_0=0,\qquad
z_t=\eta A\sum_{j=0}^{t-1}(1-\eta\kappa)^j\rho^{t-1-j}.
\label{eq:router-transient-model}
\end{equation}
These are assumptions on the update field with loads recomputed, not derivatives through stop-gradient or consequences of the auxiliary objective alone.
For $0<\eta\kappa<1$, $z_t$ initially increases and eventually returns to zero, permitting an entropy valley and recovery.
With $\lambda=a_1=0$, instead $z_t=\eta A(1-\rho^t)/(1-\rho)$; adding persistent forcing $a_\infty\ge0$ to the stable model gives $z_\infty=a_\infty/\kappa$, allowing recovery short of uniformity.
For common $A,\kappa,\rho$, concentration at step $t\ge1$ increases with LR if $t\eta\kappa\le1$: each summand's factor $\eta(1-\eta\kappa)^j$ has nonnegative derivative, with the $j=0$ derivative positive.
This sufficient condition does not predict the observed transition time.

Perturbations satisfy $e_{t+1}=(1-\eta\kappa)e_t$, hence contract exactly when $0<\eta\kappa<2$.
The valley above uses $\eta\kappa<1$; the broader contraction condition illustrates a step-size limit, without identifying the empirical critical LR or proving full-model divergence.

\subsection{GBS, Margin Fluctuations, and Entropy}
\label{app:router-finite-step}
\label{app:router-gbs}

We derive the conditional variance law used in \emph{GBS Controls Margin Fluctuation} (Section~\ref{sec:router}) and its local entropy consequence.

\paragraph{Conditional margin variance.}
For a fixed probe and expert pair, let GBS $B$ count sequences of fixed length $L$, so $M=BL$, and write $\widehat\gamma_{h,B}=B^{-1}\sum_bY_{h,b,B}$, where $Y_{h,b,B}$ is the negative margin-gradient sum within sequence $b$, divided by $L$.
For this total task--auxiliary feedback estimate (Equation~\ref{eq:router-margin-update}), define at a common fixed state $\gamma_{h,B}=\mathbb E_t\widehat\gamma_{h,B}$ and $\xi_{h,B}=\widehat\gamma_{h,B}-\gamma_{h,B}$, so $\mathbb E_t\xi_{h,B}=0$.
Then
{
\begin{equation}
\begin{aligned}
\operatorname{Var}_t(\Delta m_{j^\star h})
=\eta^2\operatorname{Var}_t(\xi_{h,B})
=\frac{\eta^2}{B^2}\sum_{b,c}\operatorname{Cov}_t(Y_{h,b,B},Y_{h,c,B})=\frac{\eta^2\sigma_h^2}{B}.
\end{aligned}
\label{eq:router-gbs-law}
\end{equation}
}
The last equality requires conditionally iid sequence contributions with a law unchanged by $B$, or conditionally iid balancing groups of fixed size and law, with $B$ divisible by that size.
For fixed groups, within-group dependence is absorbed into $\sigma_h^2$; token independence within sequences is unnecessary.
The total coefficient includes task--auxiliary covariance.
Global-batch load aggregation can change both the mean feedback (whose $B$ subscript is suppressed in the main text) and covariance, so the covariance sum is the general expression.
This margin-variance law needs no near-uniform assumption; excursions can receive subsequent reinforcement when net feedback is positive and the winner persists, but greater variance alone does not establish collapse.

\paragraph{Local entropy consequence.}
Assuming the same iid/fixed-group conditions for the full gradient contributions, write $\widehat G_{t,B}=G_t+\Xi_{t,B}$ with $\mathbb E_t\Xi_{t,B}=0$ and $\mathbb E_{t,x}\|\Xi_{t,B}^\top x\|_2^2=V_t/B$, including task--auxiliary covariance.
For centered weights $W_t=W_{R,c}(t)$, define mean-step logits $\bar\delta=(W_t-\eta G_t)^\top x$ and noise $\epsilon=-\eta\Xi_{t,B}^\top x$.
Taylor expansion at $\bar\delta$ gives
{
\begin{equation}
\begin{aligned}
\mathbb E_{t,x}H(\operatorname{softmax}(\bar\delta+\epsilon))
&=\mathbb E_xH(\operatorname{softmax}(\bar\delta))
-\frac{\eta^2V_t}{2nB}+\mathcal E_{t,B},\\
|\mathcal E_{t,B}|&\le C\,\mathbb E_{t,x}
\left[\|\bar\delta\|_2\|\epsilon\|_2^2+\|\epsilon\|_2^3\right].
\end{aligned}
\label{eq:entropy-noise-expansion}
\end{equation}
}

The linear term vanishes by conditional centering, and the entropy Hessian on centered logits is $-C_n/n+O(\|\bar\delta\|_2)$.
Both mean-step and realized logits must remain in a fixed small neighborhood, with finite displayed moments.
The leading noise-induced entropy decrease scales as $\eta^2/B$, relative to the same mean update.
A batch-size comparison additionally needs a common mean update and control of the remainder; this is not an ordering of entire training trajectories.

\begin{figure}[H]
\centering
\includegraphics[width=0.98\textwidth]{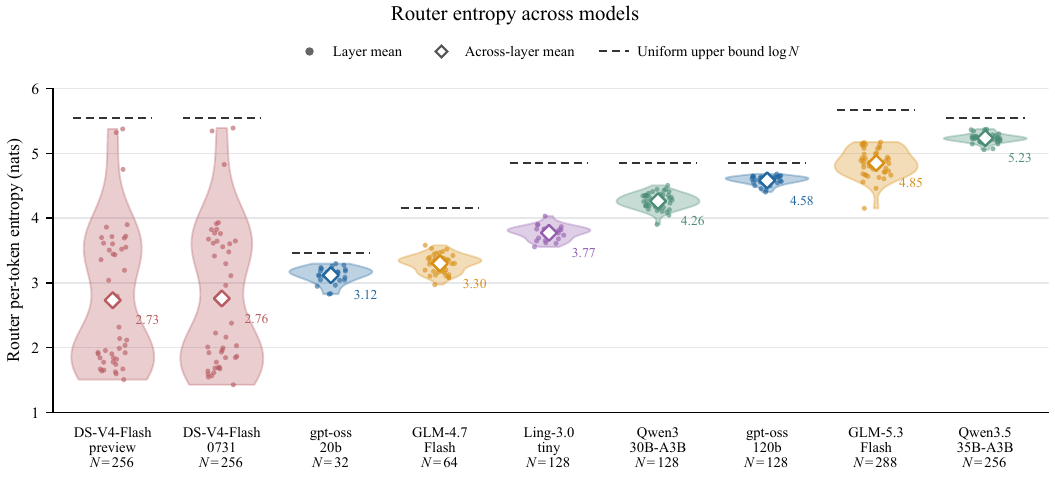}
\captionsetup{labelfont={color=black},textfont={color=black}}
\caption{Layer-wise mean per-token routing entropy across checkpoints. We apply softmax to all expert logits and average token entropy within each layer on 64 English WildBench v1 queries during prefill. Dots denote layers, violins their distributions, and diamonds model means; $n$ denotes the number of experts.}
\label{fig:router-entropy-openmodels}
\end{figure}

\begingroup
\color{black}
\subsection{Entropy Across Open-Source Checkpoints}
\label{app:router-openmodels}
Figure~\ref{fig:router-entropy-openmodels} shows layer-wise routing entropy on 64 WildBench~\citep{lin2024wildbench0} prompts. Most models' routing scores are distributed close to the uniform limit (black dashed lines).
GPT-OSS-120B and Qwen3.5-35B-A3B both have normalized mean entropy of approximately $94.4\%$, despite markedly different weight similarities ($\sim\!0$ v.s. $0.505$).
Conversely, GLM-5.3-Flash and DeepSeek-V4-Flash-preview have relatively similar weight similarities ($0.163$ v.s. $0.180$) but different normalized entropies ($85.7\%$ v.s. $49.3\%$), showing that similarity alone does not consistently describe cross-model routing concentration.
We also note that most models show consistent routing behavior across layers, with only mild depth trends. The DeepSeek-V4-Flash family is a striking exception: per-token routing entropy approaches the uniform limit only in the first few MoE layers, then decreases markedly with depth, collapsing to a narrow effective expert pool in the deepest layers. It remains an open question whether it is a stable design characteristic of the family or a training artifact.
\par
\endgroup

\section{Exact Low-Rank Spectral Computation and Monitoring Cost}
\label{app:monitoring-cost}

Section~\ref{sec:flash} describes the exact small-core computation of
the nonzero singular spectra of $\Done$, $\Dtwo$, and $\Dthree$.
Here we benchmark its cost and numerical accuracy against full-matrix
computation on a single Ascend 910B NPU.
The compressed-core computation gives large speedups at realistic
hidden sizes while preserving the spectrum to small relative error.

\begin{table}[htbp]
\centering
\small
\setlength{\tabcolsep}{4pt}
\begin{tabular}{rcccc}
\toprule
Size & Full eigvalsh (ms) & Core+eig. (ms) & Speedup & Rel. diff \\
\midrule
100  & 0.55    & 3.80   & 0.14x  & $9.89{\times}10^{-6}$ \\
200  & 39.23   & 4.54   & 8.64x  & $2.78{\times}10^{-7}$ \\
500  & 66.04   & 3.86   & 17.12x & $2.52{\times}10^{-7}$ \\
1000 & 166.37  & 3.38   & 49.18x & $2.50{\times}10^{-7}$ \\
2000 & 1512.59 & 30.09  & 50.26x & $2.29{\times}10^{-7}$ \\
4000 & 3602.54 & 124.01 & 29.05x & $2.26{\times}10^{-7}$ \\
\bottomrule
\end{tabular}
\caption{Single Ascend 910B NPU timing for full-matrix eigendecomposition versus compressed-core computation of the same singular spectrum.}
\label{tab:qk-low-rank-svd-speedup}
\end{table}

\end{document}